\pdfoutput=1

\documentclass[11pt]{article}

\usepackage{acl}

\usepackage{times}
\usepackage{latexsym}

\usepackage[T1]{fontenc}

\usepackage[utf8]{inputenc}

\usepackage{microtype}
\usepackage{xcolor}
\usepackage{bm}
\usepackage{array}

\usepackage{amssymb}
\usepackage{amsmath}
\usepackage{graphicx}
\usepackage{subfigure}

\usepackage{booktabs}
\usepackage{multirow}
\usepackage{marvosym}
\usepackage{stfloats}
\usepackage{float}

\usepackage{inconsolata}

\title{Sharing, Teaching and Aligning: Knowledgeable Transfer Learning for Cross-Lingual Machine Reading Comprehension}

\author{
Tingfeng Cao$^{1,2,3}$\thanks{\ \ Work done during an internship at Alibaba.},\ \ 
Chengyu Wang$^{2\dagger}$,\ \ 
Chuanqi Tan$^{2}$,\ \ 
Jun Huang$^{2}$,\ \ 
Jinhui Zhu$^{1,3}$\thanks{\ \ C. Wang and J. Zhu are co-corresponding authors.}\\
$^1$School of Software Engineering, South China University of Technology, China\\
$^2$Alibaba Group, China\\
$^3$Key Laboratory of Big Data and Intelligent Robot (South China University of Technology) \\ Ministry of Education, China\\
\normalsize
\texttt{setingfengcao@mail.scut.edu.cn, csjhzhu@scut.edu.cn} \\
\normalsize
\texttt{\{chengyu.wcy, chuanqi.tcq, huangjun.hj\}@alibaba-inc.com}}

\begin{document}
\maketitle
\begin{abstract}

In cross-lingual language understanding, machine translation is often utilized to enhance the transferability of models across languages, either by translating the training data from the source language to the target, or from the target to the source to aid inference. However, in cross-lingual machine reading comprehension (MRC), it is difficult to perform a deep level of assistance to enhance cross-lingual transfer because of the variation of answer span positions in different languages. In this paper, we propose~\textbf{X-STA}, a new approach for cross-lingual MRC. Specifically, we leverage an attentive teacher to subtly transfer the answer spans of the source language to the answer output space of the target.
A Gradient-Disentangled Knowledge Sharing technique is proposed as an improved cross-attention block. In addition, we force the model to learn semantic alignments from multiple granularities and calibrate the model outputs with teacher guidance
to enhance cross-lingual transferability. Experiments on three multi-lingual MRC datasets show the effectiveness of our method, outperforming state-of-the-art approaches.
\footnote{Source codes will be publicly available in the EasyNLP framework~\cite{easynlp}. URL: \url{https://github.com/alibaba/EasyNLP}.}

\end{abstract}

\section{Introduction}

Recently, significant progress has been made in NLP by pre-trained language models (PLMs)~\cite{gpt,mbert,DBLP:conf/emnlp/ZhangDWWWLHLH22}. Yet, these models often require a sufficient amount of training data to perform well, which is difficult to achieve in cross-lingual low-resource adaptation. Although
many cross-lingual PLMs have been proposed to
learn generic feature representations \cite{mbert,xlm,xlmr,mt5,mbart}, the performance gap between source and target languages is still relatively large, especially for token-level tasks such as machine reading comprehension (MRC). 
In addition, ultra-large PLMs such as  ChatGPT~\cite{chatgpt} exhibit amazing zero-shot generation abilities over multiple languages. 
We observe that such models may not be sufficient for cross-language MRC due to 
the linguistic and cultural differences between these languages, together with the requirements of very fine-grained extraction of answer spans.

\begin{figure}[t]
\centering
\includegraphics[width=0.5\textwidth, keepaspectratio]{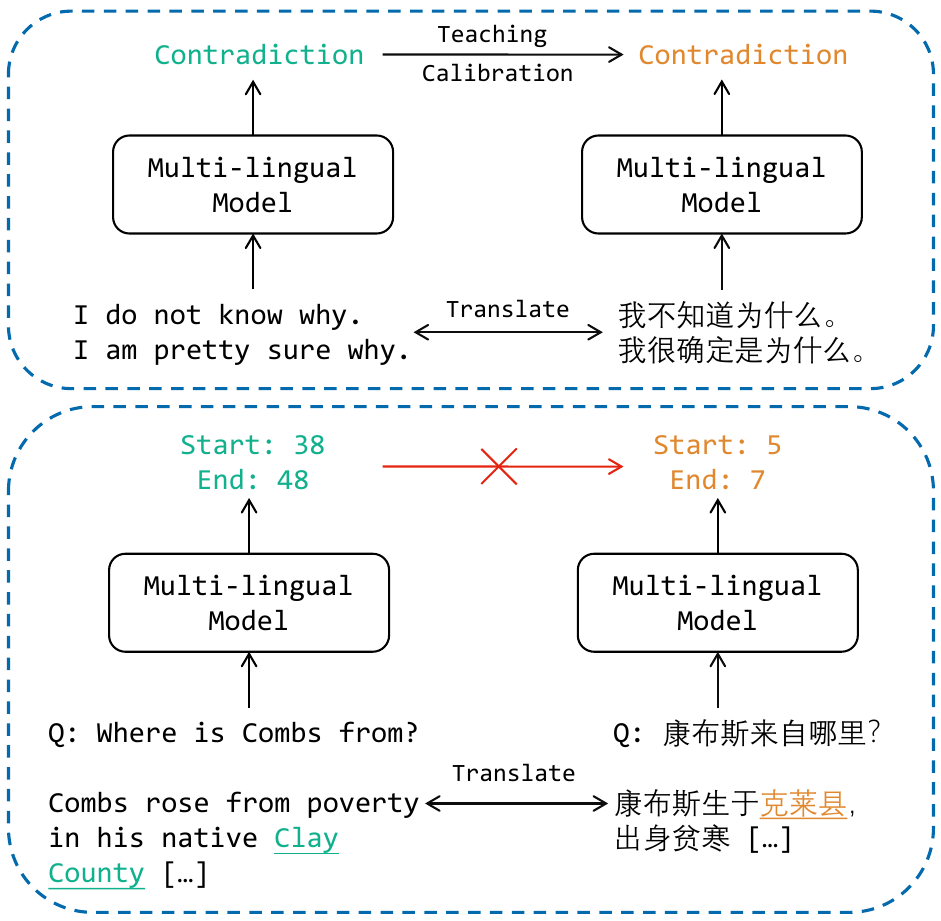}
\caption{Machine translation as an aid for cross-lingual transfer. Above is a natural language inference (NLI) task. The probability distribution of the source language can be fitted by KL Divergence for teaching low-resource languages; during inference, the target language can be translated into the source language with its output used for calibration. Below is an MRC task, where the knowledge is difficult to transfer directly.}
\label{teaching}
\end{figure}

One of the most significant challenges in cross-lingual MRC is the lack of annotated datasets in low-resource languages, which are difficult to obtain. As seen, most of the current MRC datasets are in English \cite{squad}. 
Another challenge is the linguistic and cultural variations that exist across different languages, which exhibit different sentence structures, word orders and morphological features. For instance, languages such as Japanese, Chinese, Hindi and Arabic have different writing systems and a more complicated grammatical system than English, making it challenging for MRC models to comprehend the texts. 

In the literature, machine translation-based data augmentation is often employed to translate the dataset of the source language into each target language for model training \cite{xnli, xtreme, xtreme-r}. As shown in Figure \ref{teaching}, it is relatively easy to enhance cross-lingual transferability of simple sequential classification tasks by directly fitting the output probability distribution of the source language via Kullback-Leibler Divergence \cite{filter,xtune,xmixup}. However, for MRC, it is not possible to use the output distribution of the source language directly to teach the target language, due to the answer span shift caused by translation.

Motivated by this, we propose \textbf{X-STA}, a new approach for cross-lingual MRC that follows three principles: \textbf{Sharing}, \textbf{Teaching} and \textbf{Aligning}.
For sharing, we propose the Gradient-Disentangled Knowledge Sharing (GDKS) technique, which uses parallel language pairs as model inputs and extracts knowledge from the source language. It enhances the understanding of the target language while avoiding degradation of the source language representations. For teaching, our approach leverages an attention mechanism by finding answers span from the target language's context that are semantically similar to the source language's output answers to calibrate the output answers. For aligning, alignments at multi-granularity are utilized to further enhance the cross-lingual transferability of the MRC model. In this way, we can enhance the language understanding of the model for different languages through knowledge sharing, teacher-guided calibration and multi-granularity alignment.

In summary, the main contributions of this study are as follows:
\begin{itemize}

\item We propose \textbf{X-STA}, a new approach for cross-lingual MRC based on three principles: sharing, teaching, and aligning.

\item In \textbf{X-STA}, a Gradient-Disentangled Knowledge Sharing technique is proposed for transferring language representations. Output calibration and semantic alignments are further leveraged to enhance the cross-lingual transferability of the model.

\item Extensive experiments on three multi-lingual MRC datasets verify that our approach outperforms state-of-the-art methods. Thorough ablation studies are conducted to
understand the impact of each component of our method.
\end{itemize}

\section{Related Work}

In this section, we summarize the related work in the following three aspects.


\subsection{Pre-trained Multi-lingual Language Models}

Recent work has demonstrated that large-scale PLMs have tremendous potential for downstream tasks, as well as for multilingual representations including multilingual BERT (mBERT, \citealp{mbert}), XLM \cite{xlm}, XLM-RoBERTa \cite{xlmr}, mT5 \cite{mt5}, mBART \cite{mbart}. These models extend training sets to unlabeled multilingual corpora  and project all languages into the same semantic space, allowing for cross-lingual understanding.

\subsection{Cross-lingual Knowledge Transfer}

It aims to transfer knowledge learned from a source language to target languages. A intuitive approach is to use machine translation for data augmentation \cite{xnli,qa-translate,xtreme}. Under this setting, more transferable cross-lingual representations can be learned through feature fusion \cite{filter}, consistency regularization \cite{xtune} and mainfold mixup \cite{xmixup}. However, these works are not sufficiently exploited on translation data for MRC. Other work learns language-agnostic representations through adversarial training to explicitly decompose language-specific representations \cite{adv1,adv2,advqa}, or through normalization to implicitly preserve more generic representations across languages \cite{mean,jointalign,spacenorm}. A more intuitive idea used for alignment is contrastive learning \cite{mgc,feng2022language,zhang2023cross}, where translation pairs are positive examples and texts from other pairs as negative examples.

\subsection{Cross-lingual MRC}

\citet{LAKM} propose several auxiliary pre-training tasks for solving answer boundary problems for low-resource languages. \citet{calibrenet} introduce an unsupervised phrase boundary recovery pre-training task to further address this problem. \citet{good2best} propose a two-stage step-by-step algorithm for finding the best answer from good to best for cross-lingual MRC. \citet{wu2022learning} introduce a Siamese Semantic Disentanglement Model to disassociate semantics from syntax. Our work further focuses on finding the corresponding answers from the target language  based on better knowledge transfer and textual alignments from multiple granularities.

\section{X-STA: Proposed Approach}


In this section, we present the detailed techniques of X-STA
for cross-lingual MRC.

\subsection{Task Definition and Basic Notations}
Given the a context $C$ and a question $Q$, the MRC task is to extract a sub-sequence from context $C$ as the right answer to question $Q$. Denote the input sequence as $\mathbf{X} = \{Q, C\} \in \mathbb{R}^{N}$, where $N$ is the sequence length. We use $\textbf{p}_\text{start} \in \mathbb{R}^{N}$ and $\textbf{p}_\text{end} \in \mathbb{R}^{N}$ to denote the answer start and end position probability distributions. For the sake of simplicity, we concatenate the two together to $\textbf{p} \in \mathbb{R}^{N\times 2}$. Similarly, $\mathbf{y} \in \mathbb{R}^{N\times 2}$ represents the one-hot golden label sequence. For cross-lingual scenarios, only annotated training data from the source language $D_S^\text{Train} = \{ \mathbf{X}_S^\text{Train}, \mathbf{y}_S^\text{Train} \}$ and raw test data from the target language $D_T^\text{Test} = \{ \mathbf{X}_T^\text{Test}, \mathbf{y}_T^\text{Test} \}$ are available. $S$ and $T$ denote the source and target language. Machine translation can be used to obtain training data for the target language $D_T^\text{Train} = \{ \mathbf{X}_T^\text{Train}, \mathbf{y}_T^\text{Train} \}$ and test data for the source language $D_S^\text{Test} = \{ \mathbf{X}_S^\text{Test}, \mathbf{y}_S^\text{Test} \}$~\cite{xtreme}.
In addition, we use $\mathbf{h}^l$ to denote the hidden states of a sequence in layer $l \in L$, where $L$ is the total number of transformer layers. Thus, to predict the start position and end position of the correct answer span in $\mathbf{X}$, the probability distributions $\textbf{p}$ is induced over the entire sequence by feeding $\mathbf{h}^L$ into a linear classification layer and a softmax function: $\textbf{p} = \text{softmax}(\mathbf{W}\mathbf{h}^L + \mathbf{b})$, $\mathbf{W}$ and $\mathbf{b}$ are the weights and bias of the linear classifier.


\begin{figure}[t]
\centering
\includegraphics[width=.5\textwidth, keepaspectratio]
{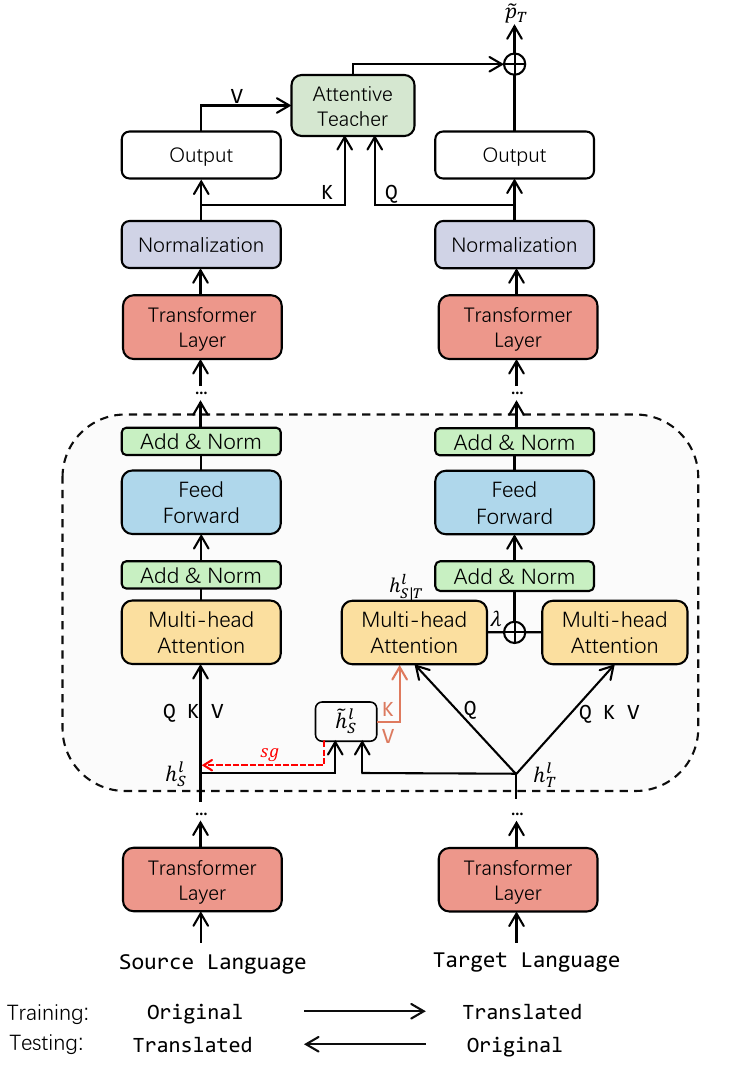}
\caption{Model architecture. The cross-attention block (GDKS) is implemented only in a certain layer. In other layers, vanilla transformer layers are applied.}
\label{model}
\end{figure}

\begin{figure}
\centering
\includegraphics[width=0.5\textwidth, keepaspectratio]{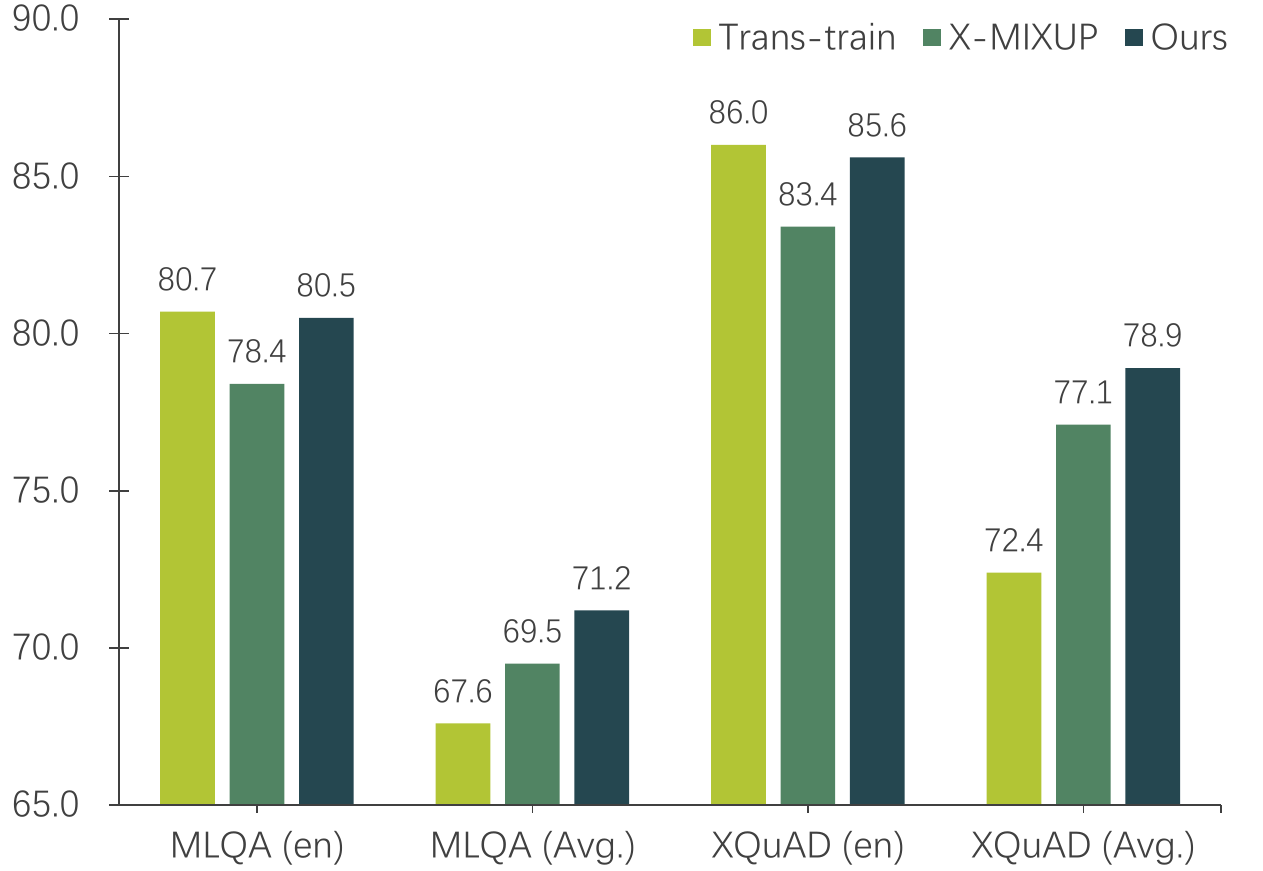}
\caption{The performance of previous methods and our method on cross-lingual MRC. X-MIXUP (a cross-attention based approach) improves the performance on target languages, but with a performance drop on the source language (en). Our approach addresses the issue by GDKS.
}
\label{en_drop}
\end{figure}

\subsection{Gradient-Disentangled Knowledge Sharing}
Although machine translation from high-resource languages to low-resource ones can be used for training multi-lingual models, the drawbacks are evident. i) Machine translation quality varies across languages. ii) The original semantics can be easily lost during translation. iii) Task labels are relatively expensive to obtain, especially for token-level cross-lingual tasks.
Thus, as shown in \ref{model}, we leverage parallel language pairs as the input and fuse cross-lingual representations. 

As in \citet{xmixup}, cross-attention can be leveraged for feature fusion.
However, a performance loss can be observed in the source language, as shown in Figure \ref{en_drop}. A reasonable conjecture is that helping the target language to extract target-related information from the hidden states of the source language leads to a degeneration of source language representations. 
To alleviate this problem, we propose Gradient-Disentangled Knowledge
Sharing (GDKS), which is an improved version of the cross-attention block. Specifically, we block the gradients from the target language output back to the source language hidden states $\mathbf{h}_S^l$. As a compensation, we add a trainable correction term:
\begin{equation*}
\tilde{\mathbf{h}_S^l} = \left\{  
             \begin{array}{ll}  
             \mathbf{h}_S^l & S = T \\
             sg(\mathbf{h}_S^l) + f(sg(\mathbf{h}_S^l), sg(\mathbf{h}_T^l)) & \text{otherwise} \\
             \end{array}  
\right.
\end{equation*}
Here, $sg(\cdot)$
is used to stop back-propagating gradients, preventing interfering with source language representations. $f(\cdot)$ refers to a trainable linear transformation with dropout. Then use the target hidden states as the query and the converted source hidden states $\tilde{h_S^l}$ as key and value to perform cross-attention, defined as follows:
\begin{equation*}
\mathbf{h}_{T\mid S}^l = \text{MHA}(\mathbf{h}_T^l,\tilde{\mathbf{h}_S^l},\tilde{\mathbf{h}_S^l}) 
\end{equation*}
where MHA is multi-head attention \cite{transformer}. Then, the target hidden states are fused with the source-aware target hidden states by the weight $\lambda$, computed as follows:
\begin{equation*}
\mathbf{h}_T^{l+1} = (1-\lambda) \cdot \mathbf{h}_{T\mid S}^l + \lambda \cdot \text{MHA}(\mathbf{h}_T^l,\mathbf{h}_T^l,\mathbf{h}_T^l)
\end{equation*}
where $\lambda=w*\lambda_0+b$, with $w$ and $b$ to be trainable parameters. 
It is worth noting that GDKS is implemented in a certain transformer layer only.

\subsection{Attentive Teacher-Guided Calibration}

As GDKS focuses on transferring knowledge from hidden states of the teacher model (trained from the source language), we also calibrate the model output distributions with teacher guidance.

\noindent\textbf{Normalization.}
The premise of obtaining good guidance is that the representations of different languages should be~\emph{normalized}
first.
Following \citet{pires2019multilingual}, we hypothesize that the representation of a multi-lingual model is composed of language-specific and language-agnostic representations. We estimate language-specific features as the mean of the language representations and remove language-specific features by subtracting the mean to retain only the generic semantic features. The intuition behind this is that a certain language may have a large number of phenomena such as function words \cite{mean}. Therefore, the average representation of that language is prominent. Inspired by Batch Normalization \cite{batchnorm}, we transform the generic semantic representation to the standard normal distribution space:
\begin{equation*}
\tilde{\mathbf{h}} = \frac{\mathbf{h}^L - \bm{\mu}_{\beta}}{\sqrt{\bm{\sigma}_{\beta}^2 + \epsilon}}
\end{equation*}
where $\bm{\mu}_{\beta}$ and $\bm{\sigma}_{\beta}$ are mean and variance of token-level representations in batch $\beta$. $\epsilon$ is a constant for numerical stability. To facilitate its use in inference,
it is set to be linguistically independent.

\noindent\textbf{Calibration.}
After normalization, we use the hidden states of the target language as query, and the hidden states and the output distribution of the source language as key and value, respectively. We also leverage MHA and average the results of the transformation of multiple heads. 
Hence, the transferred output distribution $\mathbf{p}_{_{T\mid S}} \in \mathbb{R}^{N\times 2}$ is: 
\begin{equation*}
\mathbf{p}_{_{T\mid S}} = \text{MHA}(\tilde{\mathbf{h}_T},\tilde{\mathbf{h}_S},sg(\mathbf{p}_{_{S}}))
\end{equation*}
where $\tilde{\mathbf{h}_T}$ and $\tilde{\mathbf{h}_S}$ are the normalized hidden states of source and target languages, respectively. 

During the model training phase, we incorporate a teacher-guided loss $\mathcal{L}_{tg}$ for the computation of $\mathbf{p}_{_{T\mid S}}$. Thus, tokens with the same semantics but in different languages can still be brought closer together by annotated data, even if their representations differ significantly. Specifically, we have the sample-wise loss $\mathcal{L}_{tg}$ defined as follows:
\begin{equation*}
\mathcal{L}_{tg} = - \sum_i^N\sum_j^2\mathbf{y}^{ij}log\mathbf{p}^{ij}_{T\mid S}.
\end{equation*}

For model inference, we leverage $\mathbf{p}_{_{T\mid S}}$ to calibrate the output for the target language by averaging the results from two output distributions, i.e.,
$
\tilde{\mathbf{p}_{_T}} = \frac{
    \mathbf{p}_{_{T\mid S}} + \ \mathbf{p}_{_T}
}{2}
$.

\subsection{Multi-Granularity Semantic Alignment}

We further enhance the knowledge transfer of our model, based on our proposed Multi-Granularity Semantic Alignment (MGSA) technique.

\noindent\textbf{Sentence-Level Alignment.}
A vanilla approach to learn alignments is from the sentence level.
Here, we employ Contrastive Learning (CL, \citealp{hadsell2006dimensionality, simclr}) to strengthen the alignment across languages: 
\begin{equation*}
\mathcal{L}_{align_{S}} = -log \frac{e^{\text{sim}(\mathbf{r}, \mathbf{r}^+)/\tau}}{e^{\text{sim}(\mathbf{r}, \mathbf{r}^+)/\tau} + \sum_i e^{\text{sim}(\mathbf{r}, \mathbf{r}^-_{i})/\tau}}
\end{equation*}
where $\mathbf{r}$ is the mean pooled sentence representation. $\mathbf{r}^+$ and $\mathbf{r}^-$ represent a positive sample from the parallel translated data and a negative example in the mini-batch, respectively. $\text{sim}(\mathbf{r}_1, \mathbf{r}_2)$ is the cosine similarity, i.e., $\text{sim}(\mathbf{r}_1, \mathbf{r}_2)=\frac{\mathbf{r}_1^\top \mathbf{r}_2}{\parallel \mathbf{r}_1 \parallel \cdot \parallel \mathbf{r}_2 \parallel} $. $\tau$ is the temperature hyper-parameter, which we set to 0.05 in default.

\noindent\textbf{Token-Level Alignment.}
In~\citet{qe, xmixup}, the entropy of the cross-attention distribution (ECA) is used to measure the quality of machine translation.
A smaller entropy of the attention distribution, i.e., more focused attention, can indicate a relatively higher translation quality~\cite{confidence}.
Similarly, ECA can also be used to represent the cross-lingual alignment quality, which we use as a penalty term for training the cross-lingual model to avoid distraction in GDKS.
The token-level alignment loss $\mathcal{L}_{align_{T}}$ can be defined as:
\begin{equation*}
\mathcal{L}_{align_{T}}= - \frac{1}{I} \sum^I_{i} \sum^J_{j} a^{ij} log a^{ij}
\end{equation*}
where $a^{ij}=\text{softmax}(\frac{\mathbf{h}_{T_{i}} \mathbf{h}_{S_{j}}^\top}{\sqrt{n}})$ represents attention weights, $n$ is the hidden size, $I$ is the number of target tokens and $J$ is the number of
source tokens. Next, the total alignment loss is summed by the two parts, with $\varsigma$ and $\eta $ to be the coefficients:
\begin{equation*}
\mathcal{L}_{align}=\varsigma\ \mathcal{L}_{align_{S}} + \eta \  \mathcal{L}_{align_{T}}.
\end{equation*}

\subsection{Final Training Objective}

In brief, the final training objective of~\textbf{X-STA} is:
\begin{equation*}
\mathcal{L}=\mathcal{L}_\text{MRC} + \gamma \mathcal{L}_{tg} + \mathcal{L}_{align}
\end{equation*}
where $\gamma$ is a factor for the teacher-guided loss $\mathcal{L}_{tg}$. $\mathcal{L}_\text{MRC}$ refers to the cross-entropy loss of the MRC task. Following \citet{xmixup}, we split the MRC loss $\mathcal{L}_\text{MRC}$ into the MRC loss of the source language and the target language with a balancing factor $\alpha$: 
\begin{equation*}
\mathcal{L}_\text{MRC} = \alpha\mathcal{L}_\text{MRC}^S + (1-\alpha)\mathcal{L}_\text{MRC}^T.
\end{equation*}

\section{Experiments}

\subsection{Datasets}
We evaluate~\textbf{X-STA} on three multi-lingual MRC datasets, namely MLQA \cite{mlqa}, XQuAD \cite{xquad} and TyDiQA \cite{tydiqa}.
\textbf{MLQA} is a benchmark dataset
consisting of over 5K extractive MRC instances in 7 languages: English (en), Arabic (ar), German (de), Spanish (es), Hindi (hi), Vietnamese (vi) and Chinese (zh).
\textbf{XQuAD} consists of a subset of 240 paragraphs and 1190 question-answer pairs from the SQuAD v1.1 \cite{squad} development set together with their professional translations into ten languages: English (en), Arabic (ar), German (de), Greek (el), Spanish (es), Hindi (hi), Russian (ru), Thai (th), Turkish (tr), Vietnamese (vi), and Chinese (zh).
\textbf{TyDiQA} covers 9 typologically diverse languages: English (en), Arabic (ar), Bengali (bn), Finnish (fi), Indonesian (id), Korean (ko), Russian (ru), Swahili (sw), Telugu (te). Follow XTREME \cite{xtreme}, we use the gold passage version of TyDiQA.

For the translated data, we employ the translate-train and translate-test data from XTREME \footnote{https://github.com/google-research/xtreme}. We use two evaluation metrics, namely exact match (EM) and macro-average F1 score (F1), following~\citet{squad,xtreme}.


\subsection{Experimental Settings}
We conduct extensive experiments based on two multi-lingual pre-trained backbones: mBERT \cite{mbert} and $\text{XLM-R}_{base}$ \cite{xlmr}. The batch size is set to 32. The learning rate is set to 3e-5, and decreases linearly with warmup. Following \citet{xmixup}, $\alpha$ is set to 0.2 and we implement GDKS in the 8th layer. We set $\lambda_0$ to 0.3 and $\epsilon$ to 1e-8. We perform grid search $\varsigma$, $\eta $ and $\gamma$ from [0.01, 0.05, 0.1, 0.5] on the validation set of MLQA, and finally set them to 0.05, 0.05 and 0.1, respectively. We save the model with the best averaged performance of all languages on the validation set for testing. Since there are no validation sets in XQuAD and TyDiQA. Following \citet{xmixup}, for the former, we use MLQA's validation set and for the latter, we use the English data as the validation set. All the experiments are implemented in PyTorch and run on a single server with NVIDIA Tesla V100 (32GB) GPUs. 

\subsection{Baselines}
We systematically compare our method with the following strong baselines: 
\begin{itemize}
\item \textbf{Zero-shot} models are trained on labeled data in the source language only, and directly evaluated on target languages.
\item \textbf{Trans-train} \cite{xtreme} translates training data in English into target languages. The model is trained on the combination of these original and translated training sets. 
\item \textbf{LAKM} \cite{LAKM} leverages a language-agnostic knowledge masking task by  knowledge phrases based on mBERT.
\item \textbf{CalibreNet} \cite{calibrenet} employs a unsupervised phrase boundary recovery pre-training task to enhance the multi-lingual boundary detection capability of $\text{XLM-R}_{base}$.
\item \textbf{AA-CL} \cite{good2best} is a two-stage step-by-step algorithm for finding the best answer for cross-lingual MRC over $\text{XLM-R}_{base}$.
\item \textbf{X-MIXUP} \cite{xmixup} is a cross-lingual manifold mixup method that learns compromised representations for target languages, which produces the state-of-the-art results for cross-lingual MRC.
\end{itemize}

\begin{table*}[htb]
\centering
\setlength{\tabcolsep}{5pt}{
\footnotesize
\begin{tabular}{lcccccccc}
\toprule
\textbf{Methods} & \textbf{en} & \textbf{ar} & \textbf{de} & \textbf{es} & \textbf{hi} & \textbf{vi} & \textbf{zh} & \textbf{Avg.} \\
\specialrule{0em}{1pt}{1pt}\hline\specialrule{0em}{1pt}{1pt}
Based on mBERT &  &  &  &  &  &  &  & \\
Zero-shot & 80.2/67.0 & 52.3/34.6 & 59.0/43.8 & 67.4/49.2 & 50.2/35.3 & 61.2/40.7 & 59.6/38.6 & 61.4/44.2 \\
 Trans-train & \textbf{80.7/67.7} & 58.9/39.0 & 66.0/51.6 & 71.3/53.7 & 62.4/45.0 & 67.9/47.6 & 66.0/43.9 & 67.6/49.8 \\
 LAKM & 80.1/66.9 & - & 64.4/49.9 & 69.5/51.5 & - & - & - & - \\
$\text{X-MIXUP}$ & - & - & - & - & - & - & - & 69.0/50.9 \\
$\text{X-MIXUP}^{*}$ & 78.4/64.9 & 63.3/43.5 & 67.5/53.6 & 72.3/55.0 & 65.8/47.5 & 72.2/51.7 & 66.6/45.6 & 69.5/51.7 \\
\textbf{Ours} & 80.5/67.6 & \textbf{64.1/43.7} & \textbf{69.2/54.6} & \textbf{74.2/56.5} & \textbf{67.6/49.7} & \textbf{73.5/52.8} & \textbf{69.2/47.7} & \textbf{71.2/53.2} \\

\specialrule{0em}{1pt}{1pt}\hline\specialrule{0em}{1pt}{1pt}
\multicolumn{2}{l}{Based on $\text{XLM-R}_{base}$} &  &  &  &  &  &  & \\
$\text{Zero-shot}^{*}$ & 79.2/66.2 & 56.2/37.2 & 61.7/46.9 & 67.4/50.0 & 61.5/44.2 & 65.6/45.2 & 62.5/39.0 & 64.9/47.0 \\
$\text{Trans-train}^{*}$ & 80.9/67.9 & 59.8/40.4 & 65.2/50.8 & 70.3/52.9 & 65.1/47.9 & 69.3/49.1 & 63.4/41.3 & 67.7/50.1 \\
CalibreNet & 79.7/66.6 & 56.1/37.8 & 61.7/47.6 & 68.0/50.8 & 60.0/43.8 & 66.9/46.6 & - & - \\
AA-CL & 80.1/66.8 & 58.5/41.3 & 64.6/49.8 & 69.0/51.2 & 62.8/46.5 & 67.9/47.2 & - & - \\
$\text{X-MIXUP}^{*}$ & 78.9/65.8 & 62.5/43.1 & 65.7/51.5 & 71.8/54.5 & 66.8/49.6 & 71.4/50.9 & 65.3/43.4 & 68.9/51.3 \\

\textbf{Ours} & \textbf{81.6/68.7} & \textbf{63.1/43.2} & \textbf{67.5/52.9} & \textbf{72.7/55.1} & \textbf{68.6/50.9} & \textbf{72.7/52.0} & \textbf{66.3/43.5} & \textbf{70.4/52.3} \\
\bottomrule
\end{tabular}
}
\normalsize
\linespread{1}
\caption{Overall evaluation (F1/EM) over the MLQA dataset. $^*$ denotes the results of our re-implementation.}
\label{mlqa_result}
\end{table*}


\linespread{1.1}
\begin{table*}[ht]
\centering
\setlength{\tabcolsep}{3.5pt}{
\footnotesize
\begin{tabular}{lcccccccccc}
\toprule
\textbf{Methods} & \textbf{en} & \textbf{ar} & \textbf{bn} & \textbf{fi} & \textbf{id} & \textbf{ko} & \textbf{ru} & \textbf{sw} & \textbf{te} & \textbf{Avg.} \\
\specialrule{0em}{1pt}{1pt}\hline\specialrule{0em}{1pt}{1pt}
\multicolumn{2}{l}{Based on mBERT}
  &  &  &  &  &  &  & \\
Zero-shot & \textbf{75.3/63.6} & 62.2/42.8 & 49.3/32.7 & 59.7/45.3 & 64.8/45.8 & 58.8/50.0 & 60.0/38.8 & 57.5/37.9 & 49.6/38.4 & 59.7/43.9 \\
Trans-train & 73.2/62.5 & 71.8/54.2 & 49.7/36.3 & 68.1/53.6 & 72.3/55.2 & 58.6/47.8 & 64.3/45.3 & 66.8/48.9 & \textbf{53.3}/40.2 & 64.2/49.3 \\
$\text{X-MIXUP}$ & - & - & - & - & - & - & - & - & - & 60.8/46.5 \\
$\text{X-MIXUP}^{*}$ & 72.5/60.7 & 70.0/52.8 & 55.1/41.6 & 65.8/50.0 & 74.1/57.7 & 62.6/52.2 & 63.0/43.3 & 67.5/49.1 & 51.2/37.5 & 64.6/49.4 \\

\textbf{Ours} & 73.9/63.4 & \textbf{72.4/54.5} & \textbf{60.9/47.8} & \textbf{69.4/55.9} & \textbf{76.2/60.9} & \textbf{64.0/52.2} & \textbf{65.2/46.0} & \textbf{71.2/54.1} & 51.2/\textbf{41.7} & \textbf{67.2/52.9}  \\

\specialrule{0em}{1pt}{1pt}\hline\specialrule{0em}{1pt}{1pt}
\multicolumn{2}{l}{Based on $\text{XLM-R}_{base}$}
  &  &  &  &  &  &  & \\
$\text{Zero-shot}^{*}$ & 66.0/53.4 & 61.1/41.6 & 37.8/23.0 & 61.4/45.7 & 72.6/55.0 & 48.1/33.0 & 59.5/35.0 & 54.7/35.9 & 37.5/25.4 & 55.4/38.7 \\
$\text{Trans-train}^{*}$ & 70.9/59.2 & 67.7/49.6 & 46.3/31.0 & 65.1/51.3 & 74.2/57.5 & 54.3/43.1 & 63.9/46.0 & 63.2/47.1 & 63.3/46.9 & 63.2/48.0 \\
$\text{X-MIXUP}^{*}$ & 68.0/54.8 & 67.7/48.8 & 50.6/33.6 & 66.5/52.6 & 72.0/55.0 & 52.7/40.6 & 64.0/45.0 & 64.0/47.5 & 60.2/43.3 & 62.9/46.8 \\

\textbf{Ours} & \textbf{71.3/59.3} & \textbf{68.6/50.8} & \textbf{56.7/40.7} & \textbf{67.6/54.1} & \textbf{77.7/62.5} & \textbf{55.7/44.6} & \textbf{64.2/46.1} & \textbf{64.6/48.5} & \textbf{70.2/52.9} & \textbf{66.3/51.0} \\


\bottomrule
\end{tabular}
}
\linespread{1}
\caption{Overall evaluation (F1/EM) over the TyDiQA dataset. $^*$ denotes the results of our re-implementation.}
\label{tydiqa_result}
\end{table*}





For Zero-shot and Trans-train, we report the results of mBERT from \citet{xtreme} and re-produce the results of $\text{XLM-R}_{base}$.
For LAKM, CalibreNet and AA-CL (which have been evaluated over part of our settings), we report the results from their original papers. As for X-MIXUP (the state-of-the-art method), in order to conduct a rigorous comparison, we report both the results from the original paper and our re-implementation. Among these methods, only Zero-shot and CalibreNet are under zero-shot setting, for the rest of the methods translate data are available.

\linespread{1.1}
\begin{table*}[ht]
\centering
\footnotesize
\begin{tabular}{lcccccc}
\toprule
\textbf{Methods} & \textbf{en} & \textbf{ar} & \textbf{de} & \textbf{el} & \textbf{es} & \textbf{hi} \\
\specialrule{0em}{1pt}{1pt}\hline\specialrule{0em}{1pt}{1pt}
Based on mBERT \\
Zero-shot & 83.5/72.2&61.5/45.1&70.6/54.0&62.6/44.9&75.5/56.9&59.2/46.0\\
Trans-train & \textbf{86.0/74.5}&71.0/54.1&78.8/63.9&74.2/56.1&82.4/66.2&71.3/56.2\\
$\text{X-MIXUP}$ & - & - & - & - & - & - \\
$\text{X-MIXUP}^{*}$ & 83.4/71.9 & 78.0/60.9 & 80.6/65.2 & 79.0/60.7 & 81.7/63.7 & 77.4/61.8 \\

\textbf{Ours} & 85.6/74.4 & \textbf{80.1/62.9} & \textbf{82.3/66.6} & \textbf{81.3/64.4} & \textbf{83.5/65.0} & \textbf{79.0/64.2} \\

\specialrule{0em}{1pt}{1pt}\hline\specialrule{0em}{1pt}{1pt}
\multicolumn{2}{l}{Based on $\text{XLM-R}_{base}$}\\

$\text{Zero-shot}^{*}$ & 83.3/72.3 & 66.3/50.3 & 75.4/59.4 & 74.4/56.6 & 76.1/58.6 & 67.4/50.5 \\
$\text{Trans-train}^{*}$ & 83.9/73.0 & 71.1/54.9 & 78.1/62.9 & 76.7/59.4 & 80.2/62.3 & 75.0/59.7\\
AA-CL & 84.1/73.1 & 66.5/50.3 & 77.9/62.5 & - & 80.0/61.7 & 73.8/58.9 \\
$\text{X-MIXUP}^{*}$ & 81.8/70.9 & 73.7/56.6 & 77.1/61.4 & 76.8/59.7 & 79.9/61.5 & 75.1/59.7 \\

\textbf{Ours} & \textbf{86.0/74.9} & \textbf{77.7/61.2} & \textbf{81.4/66.1} & \textbf{80.3/63.6} & \textbf{82.7/65.4} & \textbf{79.4/62.9}  \\
\specialrule{0em}{1pt}{1pt}\hline\specialrule{0em}{1pt}{1pt}
\textbf{Methods} & \textbf{ru} & \textbf{th} & \textbf{tr} & \textbf{vi} & \textbf{zh} & \textbf{Avg.} \\
\specialrule{0em}{1pt}{1pt}\hline\specialrule{0em}{1pt}{1pt}
Based on mBERT \\
Zero-shot & 71.3/53.3&42.7/33.5&55.4/40.1&69.5/49.6&58.0/48.3&64.5/49.4\\
Trans-train & 78.1/63.0&38.1/34.5&70.6/55.7&78.5/58.8&67.7/58.7&72.4/58.3\\
$\text{X-MIXUP}$ & - & - & - & - & - & 73.3/58.9 \\
$\text{X-MIXUP}^{*}$ & 80.1/63.9 & 61.9/55.7 & 75.0/58.2 & 80.4/61.1 & 71.2/61.8 & 77.1/62.3 \\

\textbf{Ours} & \textbf{81.8/66.2} & \textbf{65.2/59.4} & \textbf{76.8/62.2} & \textbf{82.0/63.8} & \textbf{70.2/60.3} & \textbf{78.9/64.5}  \\

\specialrule{0em}{1pt}{1pt}\hline\specialrule{0em}{1pt}{1pt}
\multicolumn{2}{l}{Based on $\text{XLM-R}_{base}$}\\
$\text{Zero-shot}^{*}$ & 74.4/58.8 & 64.6/53.4 & 67.7/51.1 & 73.7/53.6 & 61.4/52.4 & 71.3/56.1 \\
$\text{Trans-train}^{*}$ & 77.0/61.3 & 59.9/55.0 & 72.2/56.8 & 77.3/59.1 & 74.9/72.3 & 75.1/61.5\\
AA-CL & - & - & - & 77.6/57.5 & - & -\\
$\text{X-MIXUP}^{*}$ & 77.8/62.1 & 72.7/67.1 & 72.2/56.5 & 78.2/59.2 & 77.4/73.9 & 76.6/62.6 \\
\textbf{Ours} & \textbf{80.8/66.0} & \textbf{70.2/64.5} & \textbf{76.4/61.2} & \textbf{81.0/63.3} & \textbf{77.7/74.6} & \textbf{79.4/65.8} \\

\bottomrule
\end{tabular}
\normalsize
\linespread{1}
\caption{Overall evaluation (F1/EM) over the XQuAD dataset. $^*$ denotes the results of our re-implementation.}
\label{xquad_result}
\end{table*}

\linespread{1.1}
\begin{table}[t]
\centering
\setlength{\tabcolsep}{4pt}{
\small
\begin{tabular}{@{}lcc@{}}

\toprule
\specialrule{0em}{1pt}{1pt}
 \textbf{Ablation} & \textbf{MLQA} & \textbf{XQuAD} \\ 
 \specialrule{0em}{1pt}{1pt} \hline \specialrule{0em}{1pt}{1pt}
\ Ours & \textbf{71.2 / 53.2} & \textbf{78.9 / 64.5} \\
\hline \specialrule{0em}{1pt}{1pt}
\ \ \ \ w/o. GDKS & 71.1 / 53.0$^\dagger$ & 78.0 / 63.5$^\ddagger$ \\
\ \ \ \ w/o. ATGC & 70.8 / 52.6 & 77.4 / 63.1 \\
\ \ \ \ w/o. ATGC inference & 70.9 / 53.0 & 78.6 / 64.2  \\
\ \ \ \ w/o. Rep-Norm & 71.0 / 52.9 & 78.0 / 63.5 \\
\ \ \ \ w/o. $\text{alignment}_{s}$ & 71.2 / 53.0 & 78.5 / 64.0 \\
\ \ \ \ w/o. $\text{alignment}_{t}$ & 70.8 / 52.6 & 77.5 / 63.4 \\

\specialrule{0em}{1pt}{1pt}
\bottomrule
\end{tabular}
}
\normalsize
\linespread{1}
\caption{Ablation study of our method on MLQA and XQuAD. w/o. GDKS refers to vanilla cross-attention is used, Rep-Norm is Representation Normalization. $\text{alignment}_{s}$ and $\text{alignment}_{t}$ refer to sentence-level alignment and token-level alignment. $^\dagger$ and $^\ddagger$ have a performance drop of 1.1/1.0 and 1.3/1.0 on English.
}
\label{ablation-study}
\end{table}

\subsection{General Experimental Results}

As in Table \ref{mlqa_result}, based on mBERT, we achieved an average of 71.2\% F1 and 53.2\% EM in MLQA, exceeding all strong baselines. A gain of 1.7/1.5\% is obtained compared to the state-of-the-art X-MIXUP. As shown in Tables \ref{tydiqa_result} and \ref{xquad_result}, our method also consistently outperforms all the strong baselines on XQuAD and TyDiQA. Our method obtains on average 1.8/2.2 and 2.6/3.5 improvement F1/EM scores compared to X-MIXUP. 
In conclusion, based on two backbones, our method outperforms state-of-the-art methods on three datasets, showing the effectiveness and generalization of our method.
In addition, X-MIXUP significantly reduces the performance gap between the source and target languages, but also compromises performance on the source language; whereas our approach can achieve comparable performance to translate-train on English without negatively affecting the representation of the source language.

\begin{figure*}
\centering
\includegraphics[width=\textwidth, keepaspectratio]{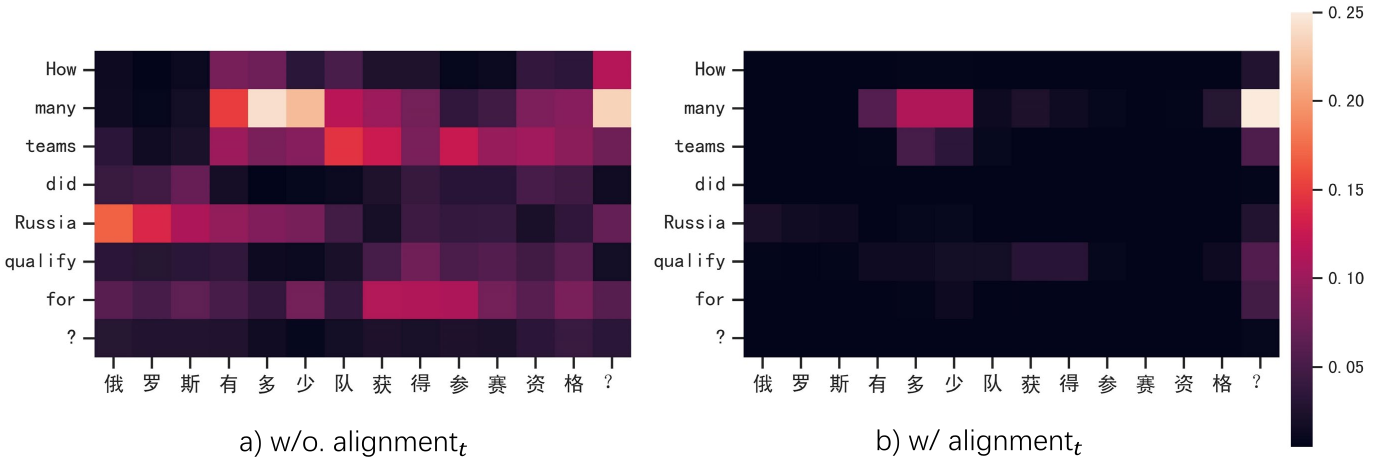}
\caption{The attention distribution heat map of query part. We show the average result of multi-head attention.}
\label{attention}
\end{figure*}

\begin{figure}[t]
\centering
\includegraphics[width=0.5\textwidth, keepaspectratio]{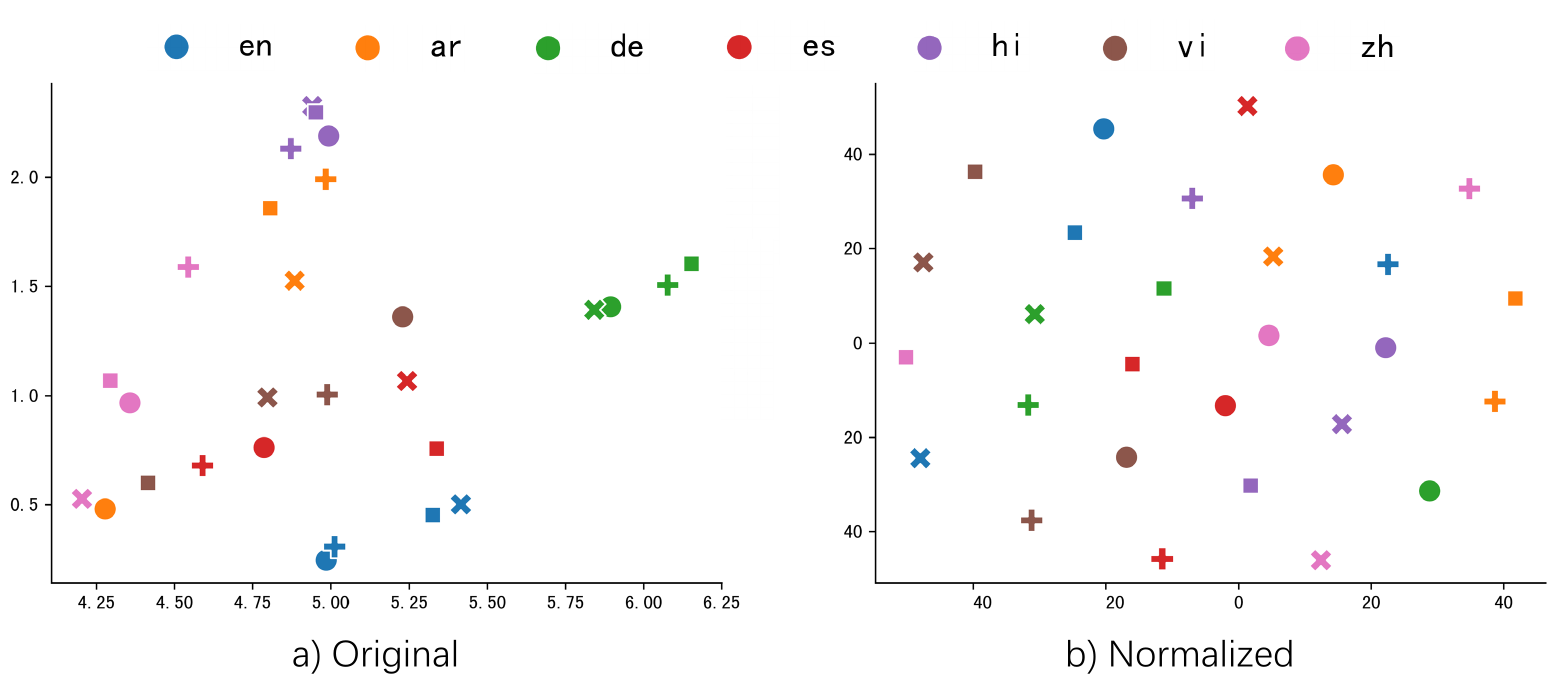}
\caption{t-SNE distributions of sentence representations. Different shapes indicate different examples.}
\label{space}
\end{figure}

\begin{figure*}
\centering
\includegraphics[width=\textwidth, keepaspectratio]{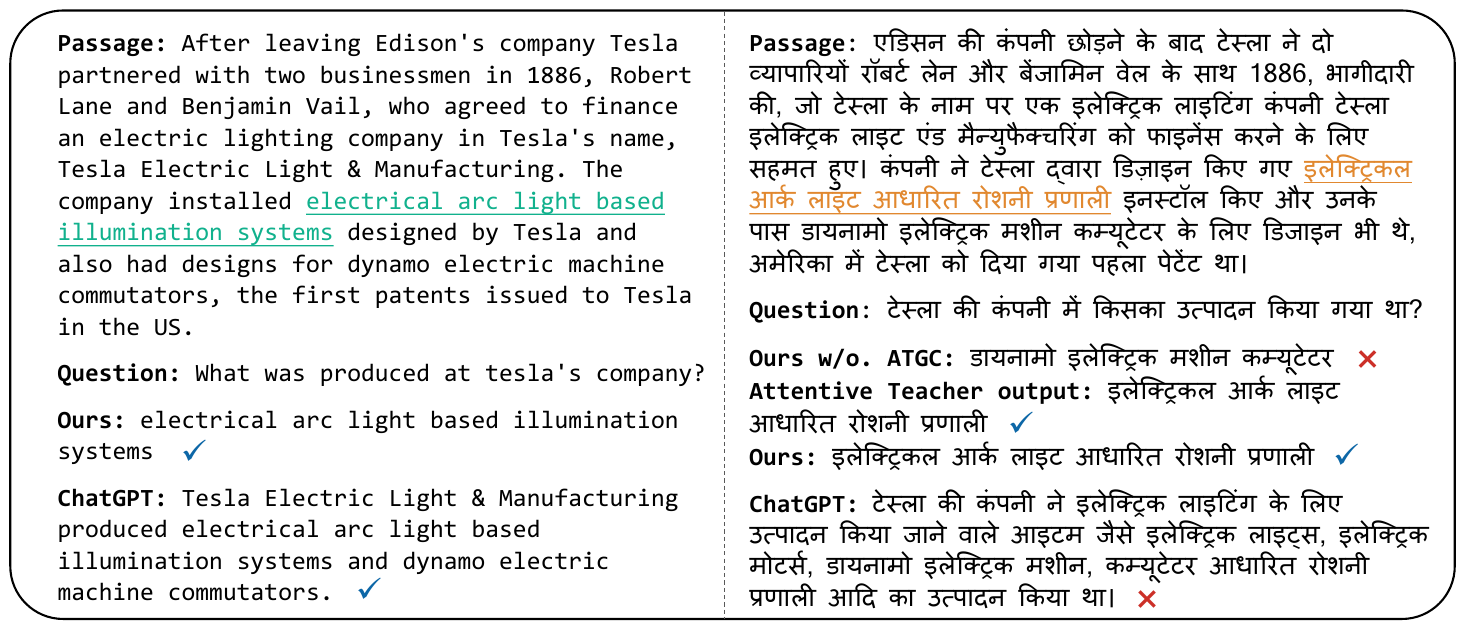}
\caption{An example from XQuAD dataset, its ground-truth answer is marked with another color and underlined. The source language example (English) on the left corresponds to the low-resource target language (Hindi) example on the right. The ChatGPT used is Mar 14 Version, and our method uses mBERT as the backbone.}
\label{case}
\end{figure*}

\subsection{Ablation Study}
We conduct an ablation study by removing each key component individually to evaluate the effectiveness of our method. As shown in Table \ref{ablation-study}, there is a performance gap when removing any of the components. 
Although the removal of GDKS has little effect on the overall performance, it significantly affects the representation of the source language, resulting in an obvious performance drop in the source language (i.e., English). 

Removing the Attentive Teacher-Guided Calibration (ATGC) component degrades the model performance the most, and the results demonstrate that mapping the output of the source language to the target language space is efficient and feasible. In addition, there is still some performance loss compared to removing ATGC only at inference time, which suggests that the improvement from ATGC does not only come from weighted outputs of source and target languages. Using the answer span as additional knowledge can enhance the cross-lingual alignment through the teacher-guided loss $\mathcal{L}_{tg}$.
Figure \ref{space} shows the visual distribution of sentence representations before and after normalization. In the original space, the distributions of the same language (same color) tend to cluster together, while after normalization, these representations are sparsely dispersed, which also shows that normalization can indeed decompose some of the language-specific representations.

Finally, we analyze the effectiveness of MGSA.
As seen, token-level alignment contributes more than sentence-level alignment. A reasonable speculation is that the MRC task is more concerned with token-level representations. As shown in Figure \ref{attention}, without token-level alignment, attention is more distracted and not well aligned across languages. Instead, token-level alignment penalizes this behavior, allowing attention to be focused on the QA-related token (e.g.,~``many'').




\subsection{Case Study}
To further demonstrate the output results of our approach, we show the answer generation process of a Hindi example and the corresponding English example from the XQuAD dataset, and compare it with a powerful ultra-large language model (i.e.,~ChatGPT). Figure \ref{case} shows that for English, both our approach and ChatGPT answer the question well. However, in a low-resource language setting such as Hindi, there are some capability limitations of mBERT and ChatGPT.
Without ATGC, our method fails to find the correct answer. When mapping the source language output to the target language output space, it successfully calibrates the output and generates the correct answer after averaging the two outputs.
ChatGPT, on the other hand, produces plausible but incorrect answers, showing a sign of producing hallucinations (also reported in~\citet{bang2023multitask}).
More cases in low-resource languages can be found in Appendix \ref{cases}.

\section{Conclusion}

In this paper, we propose~\textbf{X-STA}, which addresses the challenges of cross-lingual MRC in effectively utilizing translation data and the linguistic and cultural differences. Our work follows three principles: sharing, teaching and aligning.
Experimental results on three datasets show that our approach obtains the state-of-the-art performance compared to strong baselines. We further analyze the effectiveness of each component.
In the future, we will extend our work to other cross-lingual NLP tasks for low-resource languages.


\section*{Limitations}

Our approach requires a translation system as an aid and incurs additional inference costs during the inference process (the sequences translated back to the source language also need to go through model). 
For other cross-lingual token-level tasks (e.g., POS, NER), it is difficult to obtain the labels of translate-train data directly. Previous approaches usually use trained models to generate pseudo-labels. These low-quality labels pose significant challenges to our approach. Extending our approach to these tasks is left to our subsequent work.

\section*{Acknowledgements}

This work is partially supported by Alibaba Cloud Group, through Research Talent Program with South China University of Technology.

\newpage

\appendix

\section{Parameter Analysis}

\begin{figure}[ht]
\centering
\includegraphics[width=0.5\textwidth, keepaspectratio]{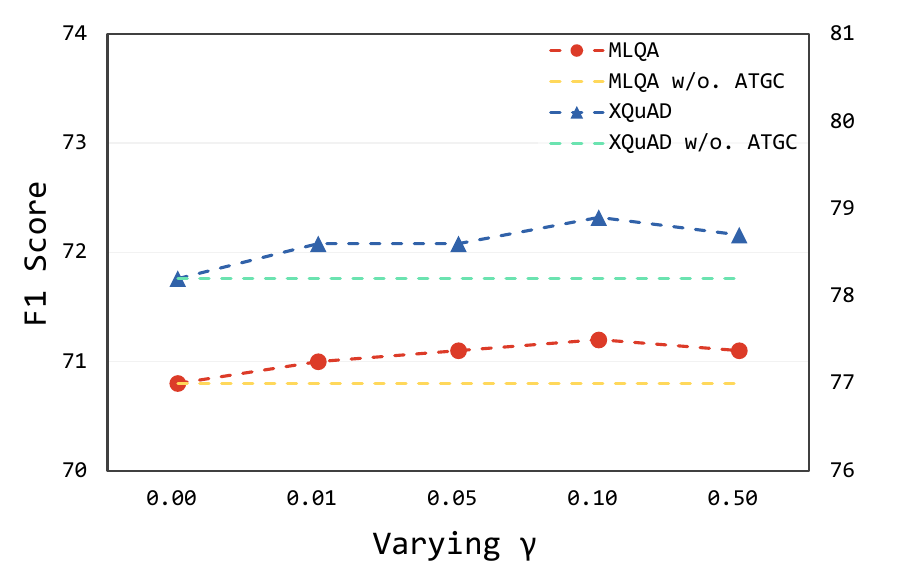}
\caption{The parameter analysis of $\gamma$ on MLQA (left vertical axis) and XQuAD (right vertical axis).}
\label{gamma}
\end{figure}

To further evaluate the effectivenss of ATGC, we conduct a series of experiments on MLQA and XQuAD for the hyper-parameter $\gamma$ = \{0, 0.01, 0.05, 0.1, 0.5\}. Figure \ref{gamma} shows that the performance trends are consistent over both datasets, with $\gamma$ achieving optimal performance on 0.1. We conjecture that when $\gamma$ is too large, it can interfere with the original representation of the model. In addition, we conduct the implementation of GDKS in different layers and find that the optimal number of layers to implement is 8, which is consistent with the results of X-MIXUP \cite{xmixup}.

\section{Cases}
\label{cases}

\begin{figure*}
\centering
\includegraphics[width=0.9\textwidth, keepaspectratio]{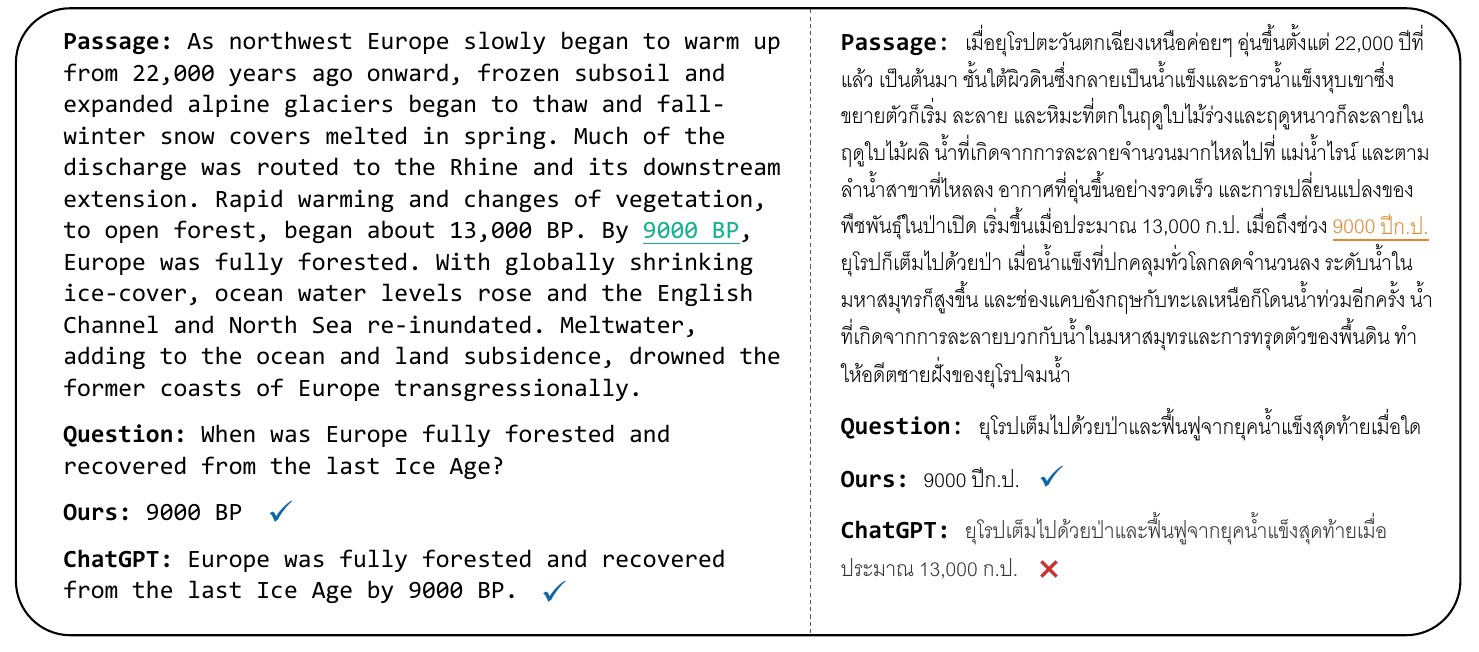}
\caption{A Thai (th) example from the XQuAD dataset.}
\label{case1}
\end{figure*}

\begin{figure*}
\centering
\includegraphics[width=0.9\textwidth, keepaspectratio]{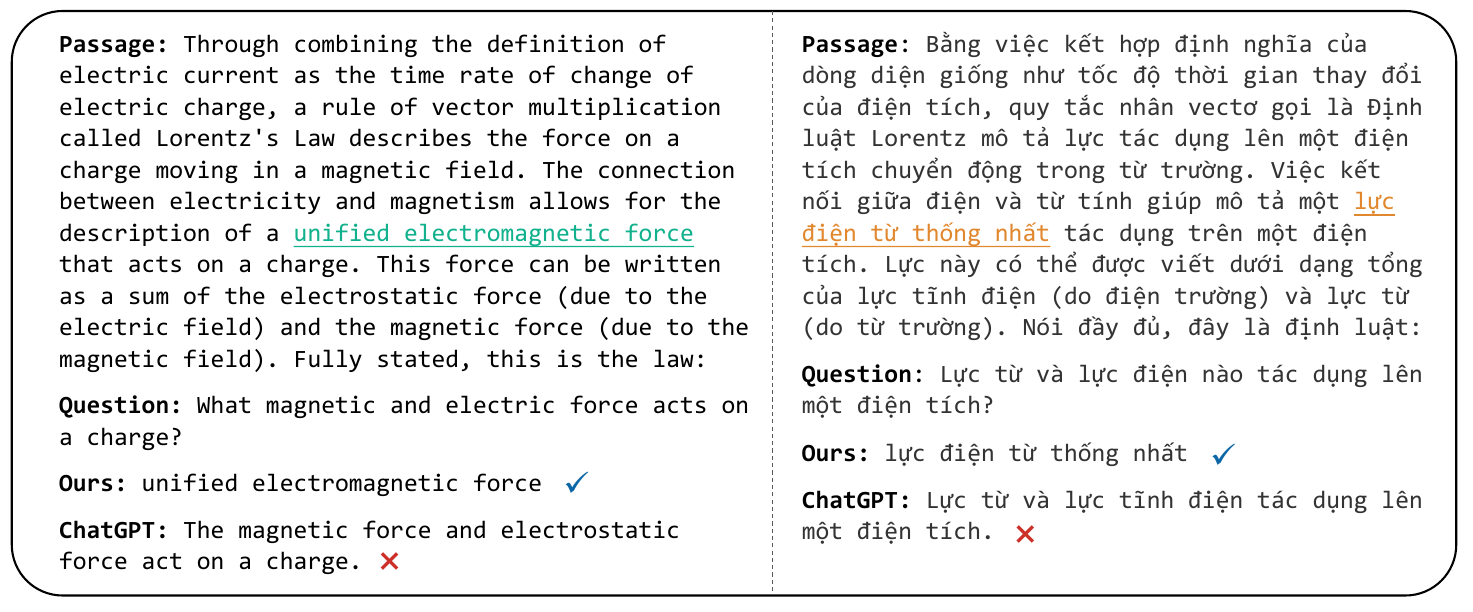}
\caption{A Vietnamese (vi) example from the XQuAD dataset.}
\label{case2}
\end{figure*}

\begin{figure*}
\centering
\includegraphics[width=0.9\textwidth, keepaspectratio]{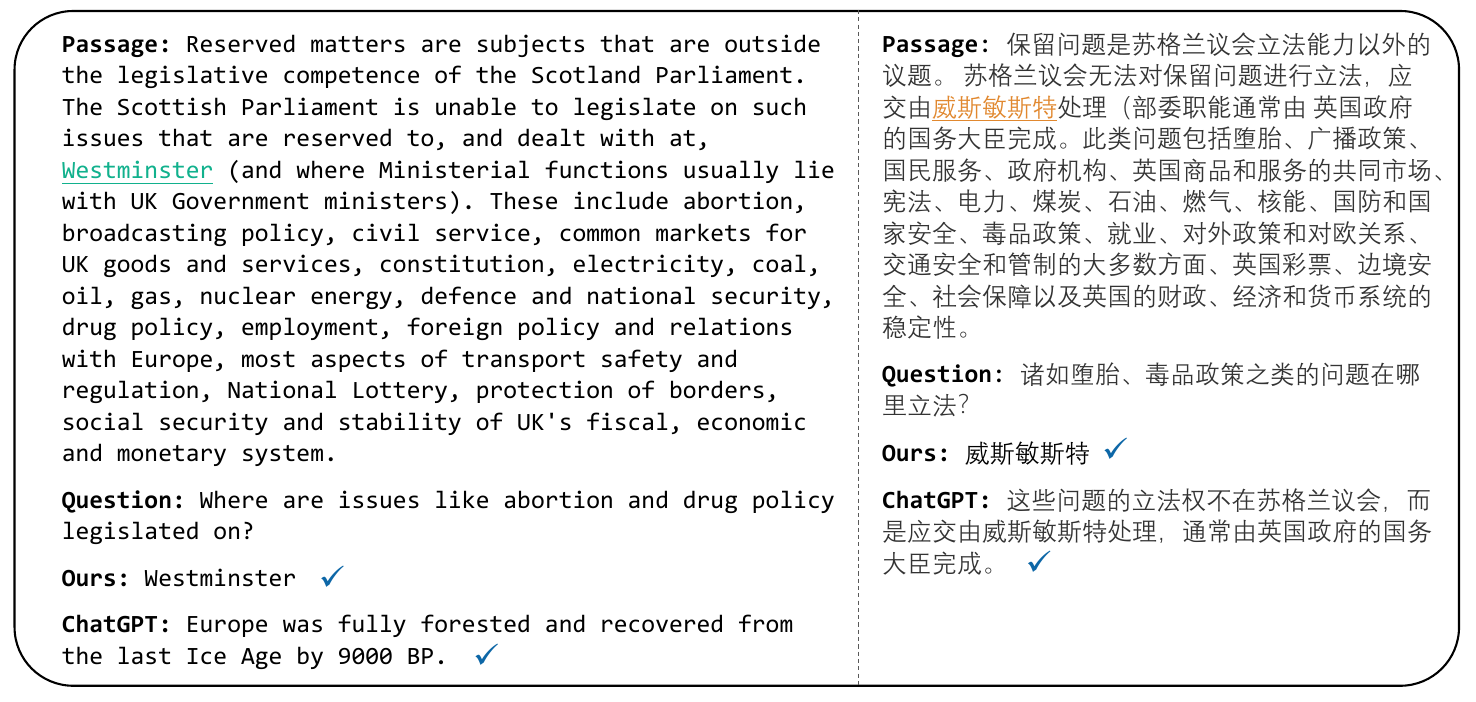}
\caption{A Chinese (zh) example from the XQuAD dataset.}
\label{case3}
\vspace{-.5em}
\end{figure*}

As shown is Figure \ref{case1}, \ref{case2} and \ref{case3}, ChatGPT sometimes 
fail to generate the answers accurately. In contrast, our method is able to extract the correct answers with the help of the English datasets.

\begin{figure}[ht]
\centering
\includegraphics[width=0.5\textwidth, keepaspectratio]{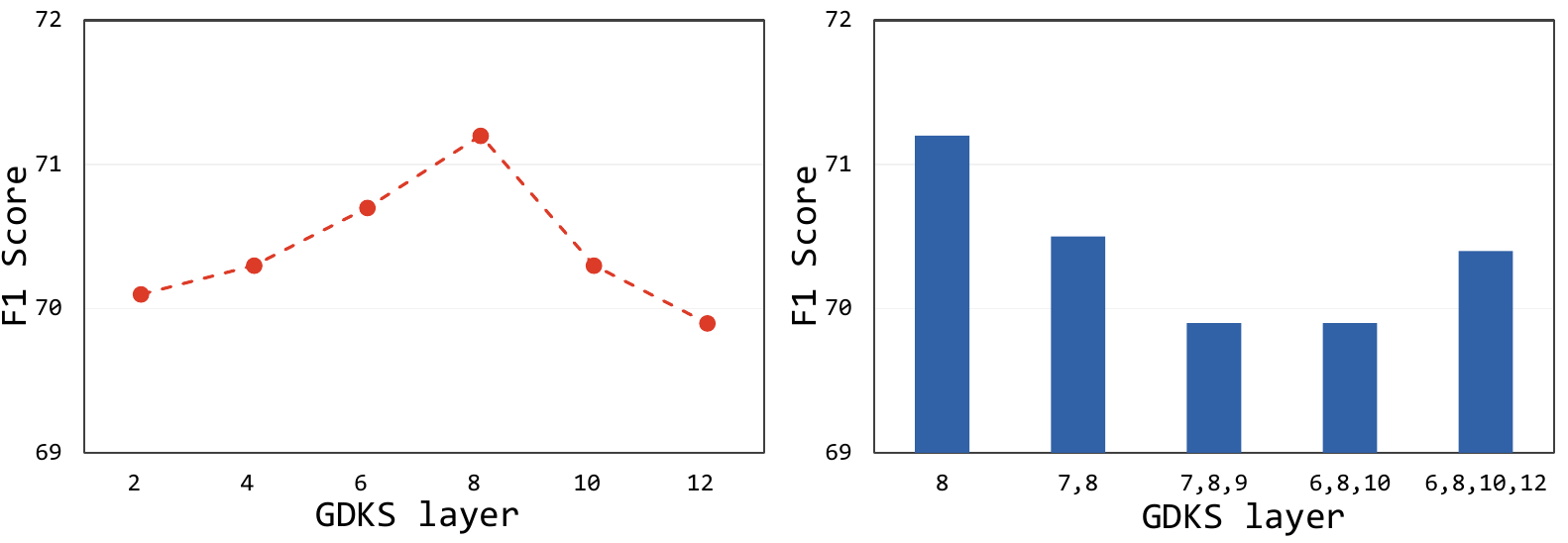}
\caption{Performance comparison on implementing GDKS in different layers over the MLQA dataset.}
\label{gdks}
\end{figure}

\end{document}